# An evolutionary computational based approach towards automatic image registration

S.K. Katiyar                                                                                                          P.V. Arun

Department Of Civil Engineering
MA National Institute of Technology, India

**Abstract**

Image registration is a key component of various image processing operations which involve the analysis of different image data sets. Automatic image registration domains have witnessed the application of many intelligent methodologies over the past decade; however inability to properly model object shape as well as contextual information had limited the attainable accuracy. In this paper, we propose a framework for accurate feature shape modeling and adaptive resampling using advanced techniques such as Vector Machines, Cellular Neural Network (CNN), SIFT, coreset, and Cellular Automata. CNN has found to be effective in improving feature matching as well as resampling stages of registration and complexity of the approach has been considerably reduced using corset optimization The salient features of this work are cellular neural network approach based SIFT feature point optimisation, adaptive resampling and intelligent object modelling. Developed methodology has been compared with contemporary methods using different statistical measures. Investigations over various satellite images revealed that considerable success was achieved with the approach. System has dynamically used spectral and spatial information for representing contextual knowledge using CNN-prolog approach. Methodology also illustrated to be effective in providing intelligent interpretation and adaptive resampling.

## 1. Introduction

Image registration involves the calculation of spatial geometric transforms that aligns a set of images to a common observational frame work. It is a key component of various image processing operations that involve different image data sets of the same area.

However, accuracy of registration techniques have been affected by various factors such as geometrical complexity, noise, vague boundaries, mixed pixel problems, and fine characteristics of detailed structures (Jacek et al., 2004). Registration techniques are broadly categorized as area based and feature based; among which area based approaches adopt a region specific matching to align the images. Feature based approaches match corresponding features so that the properties of an object being imaged can be addressable by the same coordinate pair. Accuracy attainable using area based approach has been limited due to increased resolution of images and this popularized the feature object based approaches (Vapnik et al., 1998). Feature based registration techniques generally constitute of feature detection, feature matching, mapping function design, image transformation and resampling.

Different existing feature based algorithms lack contextual interpretation capability and adopt computationally complex methods (Yuan, 2009; Sunil et al., 2004). Efficiency of these methods are situation and image-specific due to involvement of various parameters like spatial and spectral resolution, sensor characteristics etc. (Trinder & Li, 2003; Mnih & Hinton, 2010). Literature reveals a great deal of recent approaches towards the accuracy improvement of feature based strategies. There has been a long history for the usage of intelligent methodologies in the context of image registration. Soft computing techniques, such as neural networks, genetic algorithms, and fuzzy logic followed by probabilistic concepts such as random field variations, have been extensively applied in this context (Jun Liu & Helei, 2012; Gouveia, 2012; Ana et al, 2012; Jack, 1995, Chow, 2009; Zsolt Janko, 2006). Literature has also revealed many mutual information as well as intensity-based approaches (Stefan Klein, 2009; Viola, 1997; Plium, 2000; Knops, 2004; Cvejic, 2006; Mohanalin, 2009). Different entropy variations such as Tsallis, Renyis, and Shanon have been exploited for optimizing features matching. N-dimensional classifiers as well as random field concepts and different transformation techniques (SIFT, Wavelet etc) have also been applied for accurate registration (Hosseini & Homayoun, 2009; Kumar & Hebert, 2003; Chang & Kuo, 2006; Malviya & Bhirud, 2009; Hong & Zhang, 2008). Contextual information is a key factor for real-time detection to avoid ambiguity; knowledge-based classification approaches such as predicate calculus have been recently used in this context (Porway et al., 2008).

Line, point and curve based approaches found in literatures (Barbara & Flusser, 2003; Jian & Vemuri, 2005) can only compensate to simple differences between images and make use of one only kind of feature information where an attempt of a hybrid approach is neglected. Most of the feature based approaches adopt different image interpretation features namely tone, texture, pattern and colour for feature matching. Modelling of shape is not well exploited in this context and is a major factor in distinguishing different entities (Lindi, 2004). Our studies have found that inverse mapping of Cellular Automata (CA) using Genetic Algorithm (GA) can be efficiently used for modeling feature shapes. This approach seems to improve feature matching and hence registration accuracies as revealed from our investigation. Contextual information required for registration is usually represented using predicate calculus rule sets. However calculus approach fail to represent spatial relations effectively as it lack an image compatible form (Mitchell et al, 1993; Barbara & Flusser, 2003). Cellular automata rules can be used as an alternative as it can represent image rules more effectively (Orovas & Austin, 1998). Spectral and spatial information can be combined to increase the separation between object classes to yield higher registration accuracy (Mercier & Lennon, 2004). An adaptive kernel strategy (Srivastava, 2004) along with Support Vector Random Field (SVRF) may help in exploiting the n-dimensional classifier for achieving an unsupervised strategy. Main obstruct in the modelling of features using Cellular Neural Network (CNN) approach is increased computational complexity, which can be effectively tackled using an aproximation by Coreset. SIFT features are invariant to image scaling and rotation, and partially invariant to change in illumination and 3D camera viewpoint and are extensively used for registration. Object specific information can be used for optimizing and refining SIFT feature point set for an accurate registration.

Most of the registration techniques adopt static resampling techniques that are specific to certain feature geometry. High frequency aliasing artifacts present in imageries need to be preserved for detection of sub-pixel features. Typical non-adaptive interpolation methods such as nearest neighbor, bilinear, and cubic resampling yield decreasing degrees of high-frequency information fidelity. In general, higher is the order of interpolation, the smoother is the resampled image and lesser is the local contrast information. Our investigations revealed that registration of under sampled imagery requires information about the content of the scene being imaged for an accurate detection. Investigations revealed that CA based modeling can be effectively used for providing an intelligent hybridization of the existing interpolation strategies. Feature specific information is used to construct an adaptive resampler that adjusts itself with the image features to maintain sub pixel detection capabilities as well as accurate interpolation of large scale structures.

In this paper we present a registration methodology in which cellular neural network and SIFT along with adaptive kernel strategy are adopted to implement enhanced feature matching and adaptive resampling capabilities. Salient features of this work are intelligent feature matching, effective context representation and adaptive resampling for accurate registration. Accuracy of developed methodologies is compared with contemporary approaches using satellite images of Bhopal and Chandrapur cities in India.

## 2. Theoretical background

### 2.1 Resampling

Resampling or intensity interpolation is a critical step in image registration and prominent approaches in this context include nearest neighbor, Bilinear, Cubic, and Kaiser-Damped sinc (KD 16). NN selects the nearest pixel value where as Bilinear (BL) uses neighboring two points to compute required value. Cubic Convolution (CC) adopts a 4-point kernel based on cubic splines (Camann et al, 2010). Choice of resampling kernels depends on the intended use of the data. NN resampling does not alter the brightness values of the original image; however is not as visually appealing as other kernels due to its blocky effects in the image. Bilinear (BL) is a simple 2-point linear interpolator however it alters the actual image radiance. 16-point Kaiser-Damped Sinc (KD16) is based on a Kaiser windowed 16-point sinc function. MTF resampling kernel is only recommended for map and ortho corrected images and may introduce a slightly blocky appearance to the more homogeneous areas of imagery (Australian Geo science portal, 2012). An adaptive kernel strategy needs to combine these methods with respect to the situation to accomplish accurate interpolation.

### 2.2 SIFT

SIFT approach transforms images into a large collection of local feature vectors each of which are invariant to any scaling, rotation or translation of the image (Lowe, 2004). The algorithm applies a 4 stage filtering approach that includes scale-space extrema detection, orientation assignment, key point localisation, and key point description. Scale space detection attempts to identify similar locations as well as scales using Gaussian function as given in equation (1) where I is the image, G is the corresponding Gaussian representation and L is the smoothened image. Key point localisation adopts a Laplacian function (Z) as given in equation (2) to eliminate points having low contrast or poor localisation.

$$L(x, y, \sigma) = G(x, y, \sigma) * I(x, y) \quad --- \quad (1)$$

$$Z = \frac{d^2 L^{-1}}{dx^2} \frac{d L}{dx} \quad ---- \quad (2)$$

Consistent orientation to the key points based on local image properties. Gradient magnitude (M) and orientation (Φ) are computed for these points based on Gaussian smoothed image ( L) as given in (3), and (4) respectively.

$$M(x,y) = \sqrt{\left(L(x+1,y) - L(x-1,y)\right)^2 + \left(L(x,y+1) - L(x,y-1)\right)^2} \quad \text{--- (3)}$$

$$\Phi(x,y) = \sqrt{\left(\left(L(x+1,y) - L(x-1,y)\right)/\left(L(x,y+1) - L(x,y-1)\right)\right)^2} \quad \text{---- (4)}$$

An orientation histogram is formed to interpolate the peak locations and further key point description is created. The gradient information is rotated to line up with the orientation of the keypoint and then weighted by a Gaussian with variance of 1.5 * keypoint scale to get the required descriptors.

SIFT are adopted to distinguish the objects and are integrated with CNN for an optimized registration. Feature points can be accurately refined using CNN based approach and registration can be implemented in acceptable complexity.

*2.3 CNN*

CNN (Orovas & Austin, 1998; Mitchel et al, 1993) is an analog parallel computing paradigm defined in space and characterized by locality of connections between processing elements (cells or neurons). Cell dynamics of this continuous dynamic system may be denoted using ordinary differential equations as given in equation (5), where vector G is the gene which determines random nature.

$$X_k(t) = -X1 + f(G, Y_k, U_K) \quad \text{----} \quad (5)$$

CNN is effectively used for modelling object shape to improve feature matching and adaptive resampling. Random rules governing the shape of a feature can be identified by

207

evolving the feature from a single state using CNN. Abstract representations of objects are obtained by evolving features continuously until they can be separated from the background.

## 2.4 Multiple Attractor Cellular Automata (MACA)

MACA is a special type of CA with different local rules applied to different cells and will get converged to certain attractor states on execution (Weunsh et al, 2003; Sikdar et al, 2000). In an n-cell MACA with $2^m$ attractors, there exist m-bit positions at which attractors give pseudo-exhaustive 2m patterns. MACA is effectively used for feature interpretation. In order to identify an object, MACA is initialized with the pattern and operated for maximum (depth) number of cycles till it converges to an attractor. Psuedo Exhaustive Function (PEF) bits after convergence are extracted from attractor to identify class of a pattern and are compared with stored rules to interpret the object.

## 2.5 Mixture density kernel

Mercer Kernel functions can be viewed as a measure of similarity between two data points that are embedded in a high, possibly infinite dimensional feature space. Mixture Density Kernel (MDK) is a gram matrix that measures the number of times an ensemble of mixture density estimates agree that two points arise from same mode of probability density function (Srivastava, 2004). It is described using equation (6) where 'M' is the number of clusters and P ($C_m/X_i$) is the probability that data point '$X_i$' belongs to $C_m$.

$$K(X_i, X_j) = \frac{1}{Z(X_i,X_j)} \sum_{m=1}^{M} \sum_{cm=1}^{Cm} P_m\left(C_m/X_i\right) P_m\left(C_m/X_j\right) \quad ---- \quad (6)$$

Mixture density kernels are used to integrate an adaptive kernel strategy to the SVRF based clustering. This strategy facilitates learning of kernels directly from image data instead of using static kernels.

## 2.6 Support Vector Random Field (SVRF)

SVRF (Schnitzspan et al, 2008; Lee, 2005) is a Discrete Random Field (DRF) based extension for SVM. It considers interactions in the labels of adjacent data points while

208

preserving same appealing generalization properties as SVM. SVRF function is presented in equation (7), where $\Gamma_i(X)$ is a function that computes features from observations $X$ for location $i$, $O(y_i, i(X))$ is an SVM-based observation matching potential and $V(y_i, y_j, X)$ is a (modified) DRF pair wise potential.

$$P(Y|X) = \frac{1}{Z}\exp\left\{\sum_{i \in S}\log(O(y_i, \Gamma_i(X))) + \sum_{i \in S}\sum_{j \in N_i}V(y_i, y_j, X)\right\} \quad ---- \quad (7)$$

SVRF is used to implement initial clustering to segment various objects for an accurate detection and interpretation.

### 2.7 Coreset

Coreset (Agarwal et al, 2001; Badoiu et al, 2002) is small subset of a point set, which is used to compute a solution that approximates solution of entire point set. Let μ be a measure function (e.g., width of a point set) from subsets of $R^d$ to non-negative reals $R^+U\{0\}$ that is monotone, i.e., for P1 $C$ P2, μ(P1) ≤ μ(P2). Given a parameter $\varepsilon > 0$, we call a subset Q $C$ P as an $\varepsilon$ -Coreset of P (with respect to μ) if $(1 - \varepsilon)$ μ (P) ≤ μ (Q). Coreset optimisation can be adopted to reduce the number of pixels required to represent an object by preserving its shape. Hence it can be used to reduce the complexity of CA based inverse evolution.

## 3. Experiment
### 3.1 Dataset Description

Different satellite images of Bhopal and Chandrapur were used as test images for comparing the performance of various algorithms. The details of the satellite data used for these investigations are summarized in (Table 1). The ground truthing information was collected using DGPS survey conducted over Bhopal and Chandrapur during October & November 2012 respectively.

Table1. Details of dataset

| S.No | Image | Satellite | Date of Procurement | Resolution(m) |
|---|---|---|---|---|
| 1 | LISS 4 | IRS-P6 | November,2012 | 5.8 |
| 2 | LISS3 | IRS-P6 | November,2012 | 23.5 |

| 4 | PAN | CARTOSAT-1 | November,2012 | 2.5 |

**3.2 Methodology**

Edges of master (reference image) and slave (image to be registered) images are detected using Canny operator and the information is used by various stages. CA along with edge information is first used to obtain an abstract object representation of the two images (details in sec 3.2.1). The images are clustered using adaptive SVRF approach and process is enhanced with the boundary information as well as abstract object representation from 3.2.1 (details in sec 3.2.2). Detected objects along with boundary information are optimized using corset approach to reduce the complexity of shape modeling (details in sec 3.2.3). The approximated objects along with edge information are used to model feature shapes using CNN&MACA. Detected objects are further interpreted using a shape-rule mapping which stores a mapping between objects and MACA rules (details in sec 3.2.4). SIFT feature vectors of both images are refined using object information. Master and slave images are matched based on corresponding features and transformation function is devised. Slave image is further transformed using the transformation function (details in sec 3.2.5) and adaptive resampling strategy is used to resample the slave image (details in sec 3.2.6). A schematic representation of adopted methodology is presented in (Figure 1).

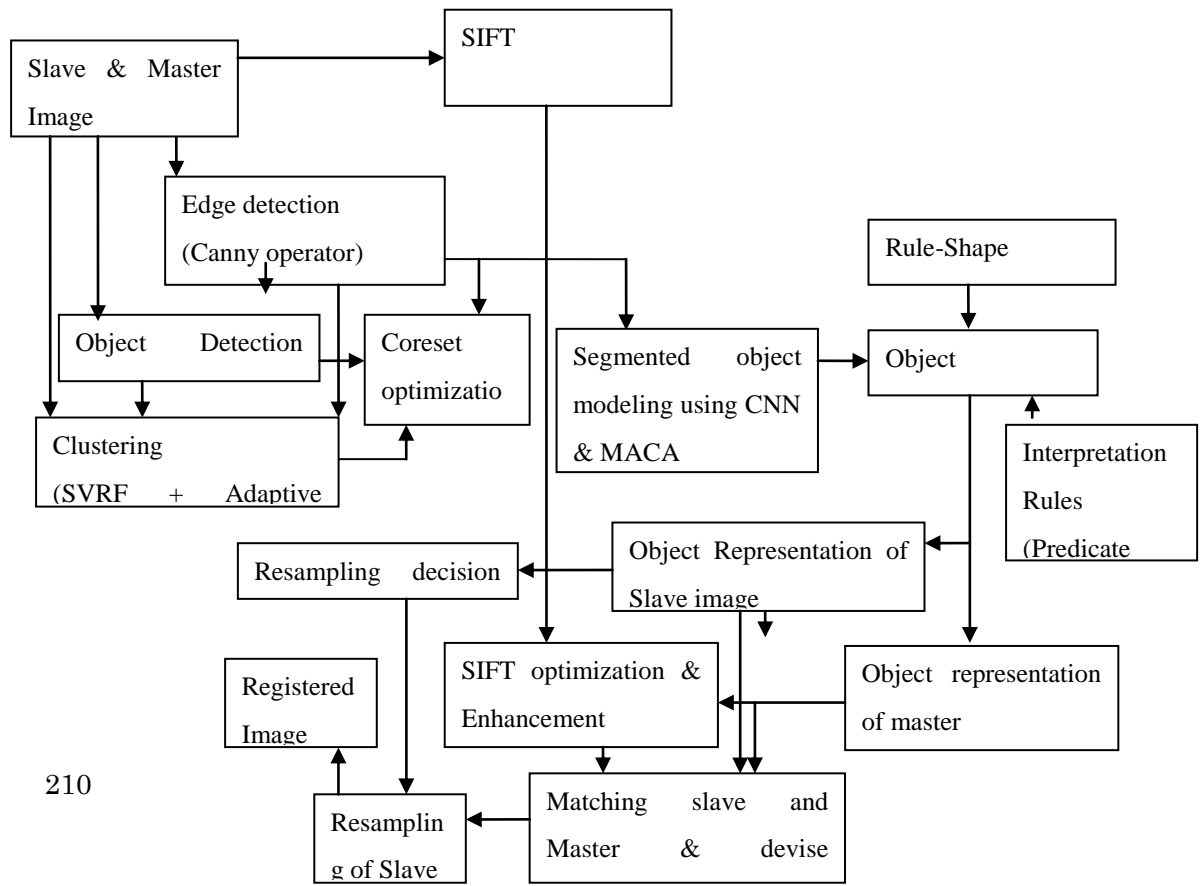

Fig. 1. Methodology work flow

*3.2.1 Object extraction*

Once the edges of slave and master images are detected using Canny operator, CA based region growing strategy is adopted to have approximate extraction of objects. Clustering is accomplished by using SVRF approach that uses a mixture density kernel. Feature specific information derived from edge information as well as abstract object approximation is used for determination of clustering parameters (seed pixel position, number of clusters), rather than adopting a random approach.

Mixture density kernels are used as it avoids the static nature of usual kernel based strategies. As shown in equation (8) kernels are learnt dynamically from the data.

$$K(X_i, X_j) = \frac{1}{Z(X_i,X_j)} \sum_{m=1}^{M} \sum_{cm=1}^{Cm} \frac{P_m^2\left(X_i,X_j/C_m\right) P_m^2(C_m)}{P_m(X_i)} \quad \text{----} \quad (8)$$

Thus parameters of mixture density kernels are adjusted automatically based on ensembles, and hence can be exploited to incoporate contextual information as well as adaptive kernel strategy for SVRF. Kernels include contextual information by incorporating both spectral and spatial information using CA rules. Preferably a weighted combination of kernels are adopted as discussed in (Hosseini & Homayoun, 2009) such that $K(P, P_i) = \mu K_x(P, P_i) + (1 - \mu)K_y(P, P_i)$. The value of tuning parameter $\mu$ is adjusted accordingly based on feature metadata using GA based approach. Edge information along with texture features are used for facilitating an accurate clustering.

*3.2.3 Coreset optimisation*

Objects in satellite images may span over a long distance and hence a reduction in pixel number (n) is needed to accomplish object modeling in acceptable complexity range. Approximation of features using a corset based approach will help to reduce the number of pixels considerably without losing original shape. For a given Feature *F*, a line corset (*k*, ε) is constructed so that it resembles *F* with much lesser number of pixels, *K<<F*.

Given any *n* points in $R^d$, a $(k, \varepsilon)$ line coreset of size polygon $\log(n)^x$ can be constructed in $nd.k^{O(1)} = O(n)$ time, given $x = 2^{O(k)} (\frac{1}{\varepsilon})^{d+2k} \log^{4k}(n)$.

The parameters such as K-line segments that are required for coreset generation are derived from edge information and abstract object representations. GA is used along with coreset to implement a shape preserved approximation. The best suitable possibilities of approximations are selected using GA that measures the similarity of shapes by comparison of PEF bits.

*3.2.4 Feature matching*

Corresponding features in both images are detected by intelligently interpreting them. CNN along with GA can be effectively used to find rules that iterates from a given initial state to a desired final state. This inverse mapping or evolution is exploited to model feature shapes in both images, and CNN rules used to evolve a particular feature is used to distinguish it. Rules corresponding to various features are thus deducted and are stored in a prolog DB. In addition to feature interpretation, these rules are also used to guide mutation and crossover of GA to increase efficiency. The inverse evolution can be attained in lesser than $n \log(n)$ time, provided that the features will converge to lower class CA configurations. Core set based optimization is used for feature approximation so that features can be effectively mapped to lower class CA configurations. MACA is automatically initialised with most likely pattern, to identify the class of the pattern in less than $\log(n)$ time.

*3.2.5 Transformation*

Once the corresponding features are modelled; master and slave images are aligned in accordance with the corresponding features. SIFT feature vectors of both images are refined using object information and are used to compute the corresponding points in both images. Once certain corresponding points in both images are determined; it is required to compute a transformation function that will determine the correspondence between the remaining points of slave and master. Different transformation functions are available over literature and prominent approaches include Thin Plate Spline (TPS), Multi Quadtratic (MQ), Weighted mean (WM) and Piecewise Linear (PL). We have adopted a weighted mean approach as suggested by Zagorchev & Goshtasby (2006) because WM is preferred over TPS, MQ, and PL when a very large set of control points with positional inaccuracies is given. WM is preferred as it uses an averaging process that smoothes the noise and does not require the solution of a very large system of equations.

*3.2.5 Adaptive Resampling*

Slave image intensities need to be interpolated over the transformed frame work and an adaptive approach is adopted to improve accuracy. Different interpolation techniques are combined at different scales based on specific image features. Feature information along with image scale is used to select the appropriate resampler. Image is transformed to Laplacian pyramid representation and is further categorized based of features using stored resampling rules. NN interpolation is adopted for small and random features in first level in the pyramid, while BL, cubic spline, or KD-16 interpolation is used for subsequent levels. Pyramid transformation is then inverted to obtain the registered image. An important consequence of this approach is that accuracy will be enhanced at various situations, especially when an object is smaller than a single pixel but exhibits high local contrast. This is because our approach gives similar results to nearest neighbor at very fine scales, but with a cubic spline interpolant's structure superimposed. This structure comes from levels farther down in the Laplacian pyramid, where the local contrast from the immediate neighborhood at every scale is effectively combined with the sub-pixel feature.

**4. Results**

Investigations of registration process over various satellite images revealed that considerable success was achieved with the approach. Different recent techniques have been compared with the developed method and accuracy has been verified on diverse image datasets. Normalized Cross Correlation Coefficient (NCCC) and Root Mean Square (RMSE) values along with execution time have been compared to evaluate the efficiency of traditional approaches with reference to proposed technique. NCCC measures the similarity between images and ranges from [0-1] with a value of unity indicating perfectly registered images. RMSE value indicates the error in registration and a least RMSE value is preferred for a perfect registration. The execution time of different approaches have been also analyzed and is categorized as high (>1 min), medium (30-60 sec), and low (<30 sec). Ground truthing have been done by means of Google earth and Differential Global Positioning System (DGPS) survey over the study areas using Trimble R3 DGPS equipment. Visual results of the method are given in (Figure 1) and

213

results of observations are as summarized in (Table 2).

Table 2: Accuracy Comparison

| S.No | TEST DATA | | TECHNIQUE | NCCC | RMSE | EXECUTION TIME | PREFERENCE | REMARKS |
|---|---|---|---|---|---|---|---|---|
| | Master Image | Slave Image | | | | | | |
| 1 | Cartosat | LISS4 | Curve | 0.59 | 4.82 | low | Presence of linear features | Enhanced by contour information |
| 2 | Cartosat | LISS4 | Surface | 0.61 | 3.85 | low | Availability of contour information | Depends on interpolation methodology |
| 3 | Cartosat | LISS4 | Moment based | 0.53 | 3.63 | low | Presence of rigid features | Matches features of same shape |
| 4 | Cartosat | LISS4 | Automatic Chip extraction | 0.78 | 0.76 | High | - | Enhanced by using SIFT features |
| 5 | Cartosat | LISS4 | Mutual Information (MI) Based | 0.67 | 1.57 | low | Intensity variations are captured and matched | Enhance by feature based approaches |
| 6 | Cartosat | LISS4 | Fourier | 0.77 | 1.25 | High | Noisy images with gray level variations at boundaries | Enhanced by assigning variable weights to IFT |
| 7 | Cartosat | LISS4 | Wavelet | 0.85 | 0.95 | High | Enhanced by Contourlet based method ,SIFT feature matching | Enhanced by time and frequency domain |
| 8 | Cartosat | LISS4 | Neural Network | 0.63 | 2.5 | High | Prior terrain knowledge | - |
| 9 | Cartosat | LISS4 | Fuzzy, Genetic | 0.58 | 1.84 | High | Presence of linear features | Combined neural approach |

| 10 | Cartosat | LISS4 | Wavelet enhanced with SIFT Feature | 0.87 | 0.61 | High | - | Enhanced by object specific matching |
|---|---|---|---|---|---|---|---|---|
| 11 | Cartosat | LISS4 | CNN-SIFT-WAVELET Method | 0.53 | 0.42 | Low | Resolution effect & noise effect are nullified | Enhanced by entropy concepts |

Results from the table indicates that CNN based approach is performing better when compared to its traditional approaches, since for all image sets (LISS4, LISS 3 & LANSAT) NCCC is more and RMSE is less in case of the proposed method.

The visual results of few features extracted by system are given below, which also reveals the accuracy of CNN approach.

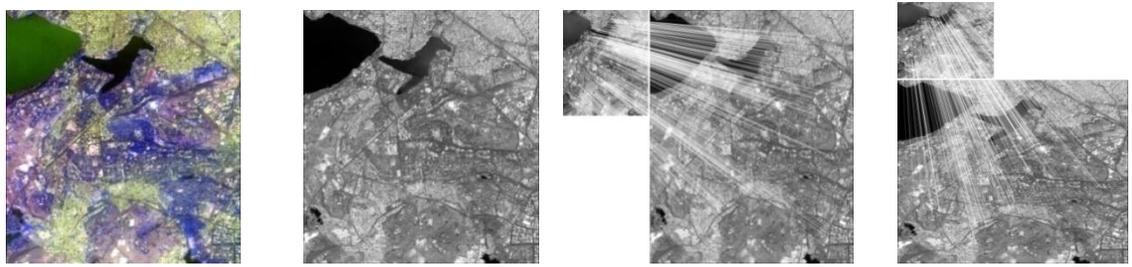

a)Slave image  b)Master image  c)Horizontal Alignment  d)Vertical Alignment

Fig.2: Registration Result

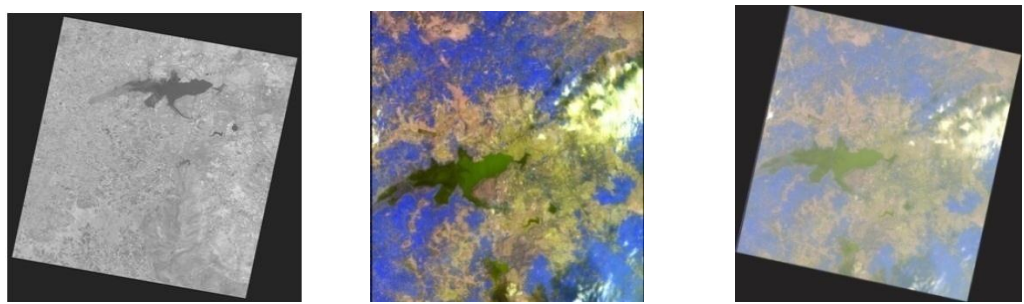

a) Master image  b) Slave image  c) Registered image

Fig 3: Registration Results

215

Detected SIFT feature points and their horizontal and vertical mapping to deduct the exact transformation is shown in (Figure 2). These figures illustrate the capability of system for effectively optimizing feature points using the contextual or interpolation rules. Results of registration of a LISS4 image with Cartosat image is given in (Figure 3). The intelligent interpretation of features have been utilized to effectively align the images and to adaptively resample without much loss of sub pixel information.

The main disadvantage of the method is its computational complexity which can be made good by corset optimization and similar approximation techniques. Complexity can be further reduced by storing the detected rule variations, and optimization methods such as genetic algorithm can be exploited to optimize the strategy. This research provides a basic framework and further investigations are needed to optimize it. Integration of fuzzy approach to the inverse resolution also seems to be promising as fuzzy / neutrosophic cognitive maps can be exploited for effectively organizing and selecting the CA rules.

## 5. Conclusion

Feature shape modeling and context knowledge representation are two important factors in distinguishing features, and lack of its consideration has hindered the accuracy of traditional feature based registration techniques. We discussed a CNN based approach that could effectively model feature shapes and contextual knowledge for accurate registration. We have illustrated the integration of these techniques to different aspects of image registration, namely feature matching and resampling. Recent random modeling and intelligent methodologies have been adopted to present a framework for the purpose. The integration of proposed framework to SIFT feature based approach for effective enhancement has been also investigated. Investigations have revealed that the method outperforms contemporary approaches in terms of accuracy and complexity. We have also suggested a new resampling method that increases the accuracy of interpolation in the regions of the images which often have sub-pixel fidelity in under sampled data.

Paper provides a framework for CA based feature shape modeling. Complexity of the approach has been considerably reduced using corset and SIFT based approximation. Proposed system has proved to be intelligent with reference to accurate registration and resampling. Disambiguation of features, enhanced detection, self learning, minimal human interpretation, and reliability are features of the system. Further investigations are needed on the improvement of proposed framework, especially on parallelizing and optimizing different operations for complexity reduction. Effective representation of different context rules also needs further improvement and techniques such as fuzzy cognitive maps seem to be promising in this context. While the proposed method shows promising results in our early experiments, there is considerable work to be done in precisely characterizing the situations in which it performs optimally.

## 6. Acknowledgement

This research work has been carried out at Maulana Azad National Institute of Technology-Bhopal, India. Authors wish to acknowledge Dr. Appukuttan K K along with other faculties of the institute for their helpful discussions and support towards this research work.

## References

Agarwal P. K., Aronov B, and Sharir M, "Exact and approximation algorithms for minimum-width cylindrical shells," *Discrete Computational Geom*etry, vol.26, no.3, pp. 307–320, 2001.

Australian Geoscience portal, available at "http://www.ga.gov.au/," Accessed on January 2012.

Badoiu M., Har-Peled S., and Indyk P., "Approximate clustering via corsets," *Proceedings of 34th Annual ACM Symposium, Theory of Computation*, pp. 250–257, 2002.

Barbara Zitová and Flusser Jane, "Image registration methods: a survey," Image Vision Computation, vol.21, no.11, pp. 977-1000, 2003.

Camann K., Thomas A., Ellis J., "Resampling considerations for registering remotely sensed images," *Proceedings of the IEEE SoutheastCon 2010 (SoutheastCon)*, pp.159-162, 18-21 March 2010.

Chang T., and Kuo C. J., "Texture analysis and classification with tree-structured wavelet transform," *IEEE Transactions on Image Processing,* vol.2, no.4, pp.429-441, 2006.

Chi K. C., "Surface Registration using a dynamic genetic algorithm," *Journal of Pattern Recognition*, vol.37, no.1, pp.105-117, 2004.

Chow C. K., "Surface Registration using a dynamic genetic algorithm," *Pattern Recognition*, vol.37, no.1, pp.105 – 117, 2009.

Cvejic N., Cangarajah C N., and Bull D. R., "Image fusion metric based on mutual information and Tsallis entropy," *Electronic Letters*, vol. 42, no.11, pp.626-627, 2006.

Fonseca L. M. G., and Costa M. H. M., "Automatic Registration of Satellite Images," *Proceedings of X<sup>th</sup> Brazilian symposium of Computer Graphic and Image Processing (SIBGRAPI),* IEEE Computer Society, Los Alamitos, California, pp. 219 – 226, 1997.

Gouveia A.R., Metz, C., Freire L., and Klein S., "3D-2D image registration by nonlinear regression," *9th IEEE International Symposium on Biomedical Imaging (ISBI)*, pp.1343-1346, 2-5 May 2012.

Hong G., and Zhang Y., "Wavelet based technique for image registration techniques for high resolution satellite imagery," *Computers and Geosciences*, vol.34, pp.1708-1720, 2008.

Hosseini R. S., and Homayoun S, "A SVMS-based hyper spectral data classification algorithm in a similarity space," *Workshop on Hyper spectral Image and Signal Processing: Evolution in Remote Sensing- WHISPERS '09*, vol.21, no.2, pp.1-4, Jun 25$^{th}$, 2009.

Jacek Grodecki, Gene Dial and James Lutes, "Mathematical Model for 3D Feature Extraction from Multiple Satellite Images Described by RPCs," *Proceedings of ASPRS 2004 Conference*, Denver, Colorado, May 23-28, 2004.

Jack J. J., "Registration of 3-D images by genetic optimization," *Pattern Recognition Letters*, vol.16, pp.823-841, 1995.

Jian Bing, and Vemuri B. C., "A Robust Algorithm for Point Set Registration Using Mixture of Gaussians," *10th IEEE International Conference on Computer Vision (ICCV 2005)*, IEEE Computer Society, Beijing, China, pp.1246-1251, 17-20 October 2005.

Jun Liu, and Helei Wu, "A New Image Registration Method Based on Frame and Gray Information," *2012 International Conference on Computer Distributed Control and Intelligent Enviromental Monitoring*, *CDCIEM-2012*, Beijing, pp.48-51, 2012.

Knops Z. F., Maintz J., Viergever M., and Pluim J., "Registration using segment intensity remapping and mutual information," *Springer MICCAI* , vol. 3216, pp.805–812, 2004.

Kumar S., and Hebert M., "Discriminative random fields: a discriminative framework for contextual interaction in classification," *Proceedings of Ninth IEEE International Conference on Computer Vision*, vol.2, no.4, pp.1150-1157, April 2003.

Lee C., Schmidt M., and Greiner R., "Support vector random fields for spatial classification," *9th European Conference on Principles and Practice of Knowledge Discovery in Databases (PKDD)*, Portugal, pp. 196, 2005.

Lindi J. Q., "A Review of Techniques for Extracting Linear Features from Imagery," *Photogrammetric Engineering & Remote Sensing*, vol.70, no.12, pp.1383–1392, December, 2004.

Lowe D. G., "Distinctive image features from scale-invariant keypoints," *International Journal of Computer Vision,* 60, 2 (2004), pp. 91-110.

Malviya A., and Bhirud S.G., "Wavelet based image registration using mutual information," *International Conference on Emerging Trends in Electronic and Photonic Devices & Systems*, ELECTRO '09,IEEE computer society, Mumbai, India pp.241-244, Aug 2009.

Mercier Gregoire and Lennon Marc, "Support Vector Machines for Hyperspectral Image Classification with Spectral-based kernels*," IEEE-Transactions of Geo Science and Remote Sensing,* Vol.45, no.3, pp. 123-130, 2003.

Mitchell M, Crutchfield J. P., and Das R, "Evolving Cellular Automata with Genetic Algorithms: A Review of Recent Work," *First International Conference on Evolutionary Computation and Its Applications (EvCA'96),* vol.1, no1, pp.120-130, May 1996.

Mnih V, and Hinton G, "Learning to detect roads in high-resolution aerial images," 11th Europian Conference on Computer Vision (ECCV), vol.10, no.1, pp.120-130, January 2010.

Mohanalin and Prem Kumar Kalra., "Mutual Information based Rigid Medical Image registration using Normalized Tsallis entropy and Type II fuzzy index," *Journal of Medical Image processing*, vol.43, no.2, pp.112-118, 2009.

Orovas C, and Austin J, "A cellular system for pattern recognition using associative neural networks," *IEEE International Workshop on Cellular Neural Networks and their Application*, vol.2, no.5, p.11, February 1998.

Porway J, Wang K, Yao B, and Zhu S. C., "A hierarchical and contextual model for aerial image understanding," *IEEE Computer Society Conference on Computer Vision and Pattern Recognition*, vol.49, no.7, pp.155-160, July 2008.

Schnitzspan P, Mario F, and Bernt S, "Hierarchical Support Vector Random Fields: Joint Training to Combine Local and Global Features," *Computer Vision – ECCV*, vol.5303, no.5, pp.527-540, May 2008.

Sikdar B. K., Paul K, Biswas G. P., Yang C, Bopanna V, Mukherjee S, and Chaudhuri P. P., "Theory and Application of GF(2P) Cellular Automata as On-Chip Test Pattern Generator," *Proceedings of 13th Int. Conf. on VLSI Design, India,* pp. 556-561, January 2000.

Srivastava A. N., "Mixture Density Mercer Kernels: A Method to Learn Kernels Directly from Data," Accepted in SIAM Data Mining Conference, 2004.

Stefan Klein, "Evaluation of Optimization Methods for Nonrigid Medical Image Registration Using Mutual Information and B Splines," *IEEE Transactions on Image Processing*, vol.16, no.12, pp.2007-2879, 2007.

Sunil R. R., Dennis D. T., Eric K, and Charles G. O., "Comparing Spectral and Object Based Approaches for Classification and Transportation Feature Extraction from High Resolution Multispectral Imagery," *ASPRS Annual Conference Proceedings*, Denver, Colorado, May 2004.

Trinder J, and Li H, "Semi-automatic feature extraction by snakes," Automatic Extraction of Man-Made Objects from Aerial and Space Images, *Birkhaeuser Verlag*, vol.25, no.1, pp.95-104, 2003.

Vapnik V., "Statistical Learning Theory," *Wiley Publishers Inc. New York*, pp.230-240, 1998.

Viola P., "Alignment by maximization of mutual information," *International Journal of Computer Vision*, vol.24, no.37, p.1541, 1997.

Yuan J, "Automatic Road Extraction from Satellite Imagery Using LEGION Networks," *Proceedings of International Joint Conference on Neural Networks*, vol.45, no.2, pp.120-130, February 2009.

Zagorchev L., and Goshtasby A, "A comparative study of transformation functions for nonrigid image registration," *IEEE Transactions on Image Processing*, vol.15, no.3, pp.529, 538, March 2006.

Zsolt Janko., Using a genetic algorithm to register an uncelebrated image pair to a 3D surface model, *Engineering Applications of Artificial Intelligence*, vol.19, no.1, pp. 269–276, 2006.